\RequirePackage{amsmath}
\documentclass[preprint]{elsarticle} 
\usepackage{setspace}
\usepackage{graphicx}
\usepackage{subcaption}
 \usepackage{caption}
\usepackage{amsfonts}
\usepackage{booktabs}
\usepackage[numbers]{natbib}
\usepackage{notoccite}
\usepackage{eucal}
\usepackage{verbatim}
\usepackage{physics}

\usepackage{lineno}

\usepackage{color}
\usepackage{tikz}
\usetikzlibrary{shapes}

\usepackage{setspace}
\usepackage{bigints}
\usepackage{siunitx}
\usepackage{lipsum}
 \usepackage{xcolor}
 \usepackage{soul}
 
\usepackage{booktabs}
\usepackage{multirow}

\usepackage[english]{babel}
\usepackage{blindtext}
\usepackage{hyperref}
\hypersetup{
	unicode=true,
	pdftitle={Deep Reinforcement Learning in Action: Experimental Control of Vortex-Induced Vibrations},
	pdfnewwindow=true,
	colorlinks=true, 	
	pdfborder={0 0 0},
	linkcolor=blue,
	linktoc=all, 		
	citecolor=blue,
	urlcolor=blue,
	breaklinks=false,
}

\usepackage{nomencl}
\makenomenclature
\renewcommand{\nomgroup}[1]{%
  \ifthenelse{\equal{#1}{A}}{\item[\textbf{Dimensional variables}]}{%
  \ifthenelse{\equal{#1}{B}}{\item[\textbf{Dimensionless parameters}]}{%
  \ifthenelse{\equal{#1}{C}}{\item[\textbf{Reinforcement learning parameters}]}}}
}

\makeatletter
\let\save@mathaccent\mathaccent
\newcommand*\if@single[3]{%
  \setbox0\hbox{${\mathaccent"0362{#1}}^H$}%
  \setbox2\hbox{${\mathaccent"0362{\kern0pt#1}}^H$}%
  \ifdim\ht0=\ht2 #3\else #2\fi
  }
\newcommand*\rel@kern[1]{\kern#1\dimexpr\macc@kerna}
\newcommand*\widebar[1]{\@ifnextchar^{{\wide@bar{#1}{0}}}{\wide@bar{#1}{1}}}
\newcommand*\wide@bar[2]{\if@single{#1}{\wide@bar@{#1}{#2}{1}}{\wide@bar@{#1}{#2}{2}}}
\newcommand*\wide@bar@[3]{%
  \begingroup
  \def\mathaccent##1##2{%
    \let\mathaccent\save@mathaccent
    \if#32 \let\macc@nucleus\first@char \fi
    \setbox\z@\hbox{$\macc@style{\macc@nucleus}_{}$}%
    \setbox\tw@\hbox{$\macc@style{\macc@nucleus}{}_{}$}%
    \dimen@\wd\tw@
    \advance\dimen@-\wd\z@
    \divide\dimen@ 3
    \@tempdima\wd\tw@
    \advance\@tempdima-\scriptspace
    \divide\@tempdima 10
    \advance\dimen@-\@tempdima
    \ifdim\dimen@>\z@ \dimen@0pt\fi
    \rel@kern{0.6}\kern-\dimen@
    \if#31
      \overline{\rel@kern{-0.6}\kern\dimen@\macc@nucleus\rel@kern{0.4}\kern\dimen@}%
      \advance\dimen@0.4\dimexpr\macc@kerna
      \let\final@kern#2%
      \ifdim\dimen@<\z@ \let\final@kern1\fi
      \if\final@kern1 \kern-\dimen@\fi
    \else
      \overline{\rel@kern{-0.6}\kern\dimen@#1}%
    \fi
  }%
  \macc@depth\@ne
  \let\math@bgroup\@empty \let\math@egroup\macc@set@skewchar
  \mathsurround\z@ \frozen@everymath{\mathgroup\macc@group\relax}%
  \macc@set@skewchar\relax
  \let\mathaccentV\macc@nested@a
  \if#31
    \macc@nested@a\relax111{#1}%
  \else
    \def\gobble@till@marker##1\endmarker{}%
    \futurelet\first@char\gobble@till@marker#1\endmarker
    \ifcat\noexpand\first@char A\else
      \def\first@char{}%
    \fi
    \macc@nested@a\relax111{\first@char}%
  \fi
  \endgroup
}
\makeatother

\begin{document}
\begin{frontmatter}

\title{Deep Reinforcement Learning in Action: Real-Time Control of Vortex-Induced Vibrations}

\author[address1]{Hussam Sababha\corref{mycorrespondingauthor}}
\cortext[mycorrespondingauthor]{Corresponding author}
\ead{haa385@nyu.edu}

\author[address3]{Bernat Font}

\author[address1,address2]{Mohammed Daqaq}

\address[address1]{Division of Engineering, NYU Abu Dhabi, Abu Dhabi, UAE}
\address[address2]{Department of Mechanical Engineering, Tandon School of Engineering, New York, USA}
\address[address3]{Department of Mechanical Engineering, Delft University of Technology, Delft, Netherlands}

\begin{abstract}
This study showcases an experimental deployment of deep reinforcement learning (DRL) for active flow control (AFC) of vortex-induced vibrations (VIV) in a circular cylinder at a high Reynolds number ($Re = 3000$) using rotary actuation. Departing from prior work that relied on low-Reynolds-number numerical simulations, this research demonstrates real-time control in a challenging experimental setting, successfully addressing practical constraints such as actuator delay. When the learning algorithm is provided with state feedback alone (displacement and velocity of the oscillating cylinder), the DRL agent learns a low-frequency rotary control strategy that achieves up to 80\% vibration suppression which leverages the traditional \textit{lock-on} phenomenon. While this level of suppression is significant, it remains below the performance achieved using high-frequency rotary actuation. The reduction in performance is attributed to actuation delays and can be mitigated by augmenting the learning algorithm with past control actions. This enables the agent to learn a high-frequency rotary control strategy that effectively modifies vortex shedding and achieves over 95\% vibration attenuation. These results demonstrate the adaptability of DRL for AFC in real-world experiments and its ability to overcome instrumental limitations such as actuation lag.
\end{abstract}


\end{frontmatter}




\nomenclature[A,01]{$M$}{Cylinder mass (kg)}
\nomenclature[A,02]{$D$}{Cylinder diameter (m)}
\nomenclature[A,03]{$K$}{Spring stiffness (N/m)}
\nomenclature[A,04]{$L$}{Immersed length (m)}
\nomenclature[A,05]{$Y$}{Transverse displacement (m)}
\nomenclature[A,06]{$A$}{Oscillation amplitude (m)}
\nomenclature[A,07]{$V$}{Free-stream flow velocity (m/s)}
\nomenclature[A,08]{$M_d$}{Displaced fluid mass (kg)}
\nomenclature[A,09]{$\rho$}{Fluid density (kg/m$^3$)}
\nomenclature[A,10]{$f_n$}{Natural frequency (Hz)}
\nomenclature[A,11]{$f_r$}{Rotational forcing frequency (Hz)} 
\nomenclature[A,12]{$\Omega$}{Cylinder rotational speed (rad/s)}
\nomenclature[A,13]{$\Omega_0$}{Reference rotational speed magnitude (rad/s)}
\nomenclature[A,14]{$T_0$}{Natural oscillation period (s)}
\nomenclature[A,15]{$\tau$}{Characteristic time scale (s)}
\nomenclature[A,16]{$T$}{Episode duration (s)}

\nomenclature[B,01]{$\alpha$}{Non-dimensional rotational speed}
\nomenclature[B,02]{$m$}{Mass ratio}
\nomenclature[B,03]{$\zeta$}{Damping ratio}
\nomenclature[B,04]{$U$}{Reduced velocity}
\nomenclature[B,05]{$SG$}{Skop–Griffin parameter}
\nomenclature[B,06]{$St$}{Strouhal number}

\nomenclature[C,01]{$s$}{State vector}
\nomenclature[C,02]{$a$}{Action }
\nomenclature[C,03]{$r$}{Reward function}
\nomenclature[C,04]{$\gamma$}{Discount factor}
\nomenclature[C,05]{$V(s)$}{State-value (advantage) function}
\nomenclature[C,06]{$Q(s,a)$}{State–action value function}
\nomenclature[C,07]{$\pi_\theta$}{Policy function parameterized by $\theta$}
\nomenclature[C,08]{$\theta$}{Neural network parameters}
\nomenclature[C,09]{$\epsilon$}{Clipping parameter for PPO}
\nomenclature[C,10]{$n$}{Number of past actions included in state history}

\section{\label{sec:intro}Introduction}
Vortex-induced vibration (VIV) of elastic bodies in cross-flow is a widely encountered and practically important fluid-structure interaction phenomenon. It arises when vortices are shed from an elastic body at a frequency close to its natural frequency, enabling effective energy transfer from the flow and into the body. This resonance phenomenon, also known as \textit{lock-in}, can incite large-amplitude oscillations, potentially compromising the performance and structural integrity of many engineering systems \citep{Williamson2004,Bourguet2011}.  Notable examples of such adverse effects include, but are not limited to, vibration of heat exchanger tubes~\cite{Shi2014}, fatigue of marine risers in offshore operations~\cite{Trim2005}, and wind-induced oscillations in tall buildings and long-span bridges~\cite{Wu2012}.

The practical significance of VIV has resulted in an extensive body of research aimed at both elucidating its underlying physics and developing strategies to mitigate it \cite{bearman2011circular,SARPKAYA2004389,Williamson2004}. This includes ongoing efforts to devise effective passive and active control techniques \cite{choi2008control,liang2018viv,Hover2001,Bearman2004}. Passive control strategies seek to influence flow separation and wake development by modifying the geometry or surface characteristics of the structure. Common approaches include the addition of helical strakes~\cite{Bearman2004}, fairings~\cite{yu2015suppression}, splitter plates~\cite{liang2018viv}, and tripping wires~\cite{Hover2001}. The main advantage of passive strategies is their simplicity and independence from external energy input, which makes them cost-effective and relatively easy to implement. However, such methods rarely eliminate VIV entirely and often introduce undesirable trade-offs, such as increased drag, which may negatively impact the overall system performance~\cite{owen2001passive}.

Given the inherent limitations of passive control methods, considerable research has also been directed toward active flow control techniques, which utilize external energy input to directly influence the vortex shedding dynamics. Such approaches aim to either weaken the strength of vortex shedding \cite{wang2016active}, or to shift the vortex shedding frequency away from the natural frequency of the structure. The latter strategy, which concerns shifting the vortex shedding frequency, was first demonstrated on a rigidly mounted cylinder, and involved forcing the cylinder to rotate at an angular speed, $\Omega$, which varies sinusoidally such that \cite{baek1998numerical},
\begin{equation}
\Omega (t)= \Omega_0 \sin(2 \pi f_r t),
\label{Omega}
\end{equation}
where $\Omega_0$ is the magnitude of the angular velocity and $f_r$ is its frequency. It was shown that vortex shedding can synchronize or \textit{lock-on}\footnote{The lock-on between the imposed sinusoidal rotary frequency and vortex shedding should not be confused with lock-in, which refers to the synchronization between the body’s natural frequency and vortex shedding} to the imposed sinusoidal rotary motion~\cite{okama1975viscous, taneda1978visual}, and that the \textit{lock-on} occurs across a wide range of forcing parameters leading to significant reduction in drag and transverse fluid forces~\cite{tokumaru1991rotary}. More recent studies have extended the implementation of \textit{lock-on} control to the suppression of VIV in elastically-mounted cylinders \cite{du2015suppression,wong2018experimental}. It was observed that, within the \textit{lock-on} regime, significant reduction in the amplitude of transverse VIV can be achieved, although notable exceptions arise when the frequency, $f_r$, of the imposed rotary motion is close to the natural frequency of the elastically-mounted cylinder, in which case large VIV amplitudes may still persist~\cite{du2015suppression}. It was also observed that the \textit{lock-on} regime widens as the magnitude of the prescribed angular velocity, $\Omega_0$, increases. Some researchers have also investigated closed-loop control strategies, in which feedback signals acquired using sensors are used to dynamically adapt the actuation of the elastic structure in a way that suppresses VIV. Compared to the more extensively studied open-loop approaches, closed-loop implementations remain relatively rare and are often limited to variants of the classical proportional--integral--derivative (PID) framework~\cite{zhang2004closed,vicente2018flow}.

In parallel with these developments, machine learning, particularly deep reinforcement learning (DRL), has recently emerged as a promising alternative to discover effective active flow control (AFC) strategies. Unlike traditional control methods,  DRL enables environment-driven learning and adaptation without requiring explicit modeling of the complex governing equations. Along this line, Rabault et al.~\cite{rabault2019artificial} reported the first application of an artificial neural network trained with a DRL agent for AFC. In their numerical study, which involved a 2D simulation of a circular cylinder at low Reynolds number ($Re = 100$), the DRL effectively learned a control policy, which manipulates the mass flow rates of two jets positioned on either side of the cylinder in such a way that stabilizes the vortex street and achieves an 8\% reduction in drag with minimal actuation effort. Since then, DRL has been successfully applied to a number of AFC problems including drag reduction \citep{Fan2020,Paris2021,Guastoni2023,suarez2025active,xia2024active, Wang2023}, collective swimmers and gliding control \citep{Verma2019,Novati2019}, separation control \citep{font2025deep,montala2025deep}, and heat transfer \citep{vasanth2024multi}, among others. The reader is referred to \citep{Vignon2023,Moslem2025} for comprehensive reviews on DRL for AFC.

Recent progress has also demonstrated the strong potential of DRL for mitigating VIV. In one numerical study, Zheng \textit{et al.}~\cite{zheng2021active} compared DRL and Gaussian-process-based active learning for controlling VIV in a circular cylinder using jet actuation at a low Reynolds number ($Re = 100$). While active learning achieved moderate vibration reduction, the DRL approach, employing a soft actor-critic algorithm, reduced vibration amplitudes by over 80\%. In another also numerical study, Chen \textit{et al.}~\cite{chen2023deep} applied DRL to suppress the VIV of a square cylinder using synthetic jets positioned on different sides of the cylinder. They achieved a vibration reduction of up to 96\% and demonstrated that DRL agents can adapt actuation strategies to actuator placement and reduced velocity conditions.  

Of particular relevance to the present study is the numerical investigation of Ren \textit{et al.}~\cite{ren2024deep}, who applied a DRL algorithm to uncover a new control strategy that imposes rotation on a circular cylinder to suppress VIV under \textit{lock-in} conditions. Using only cylinder kinematics (displacement, velocity, and acceleration), as state inputs, the DRL agent was able to successfully reduce the amplitude of VIV by up to 99.6\%.  Interestingly, the mechanism learned by the DRL differs from the well-established \textit{lock-on} effect, highlighting that DRL is capable of discovering novel strategies for attenuation. More recently, Ren \textit{et al.}~\cite{ren2024active} extended DRL-based AFC to wake-induced vibrations in tandem and staggered cylinder configurations. Using rotary actuation, the control mechanism based on the DRL agent achieved vibration reductions exceeding 95\% across various flow conditions.

While previous numerical studies have demonstrated the effectiveness of DRL for VIV suppression, these efforts have been limited to simulated environments and low Reynolds numbers, e.g. $Re\leq 300$. On one hand, simulations provide absolute control on the environment and system's state, but the high computational cost of DRL for scale-resolving simulations is currently a computational bottleneck preventing high-Reynolds-number investigations \citep{font2025deep}. As a result, the applicability of DRL-based control strategies to VIV suppression for moderate-to-high Reynolds-number flows remains unexplored. On the other hand, experimental implementation introduces practical challenges that are typically absent in simulations, including communication and actuation delays, sensor noise, and limited control bandwidth. These factors can significantly affect learning dynamics and may lead to control strategies that differ from those developed in idealized settings. Therefore, experimental validation is a critical step toward assessing the robustness and generalizability of DRL in VIV applications.

To address this gap, the present work experimentally investigates DRL-based flow control for mitigating VIV of a circular cylinder using rotary actuation. Conducting experiments at high Reynolds numbers, $Re = 3000$, enhances the relevance of the results to real-world fluid systems, which predominantly operate in turbulent regimes. Unlike earlier studies that primarily controlled the cylinder’s rotational velocity, the present approach directly modulates the actuator voltage via pulse-width modulation (PWM), thereby eliminating an intermediate control step and allowing the DRL agent to learn the voltage–rotation mapping implicitly. Beyond evaluating vibration suppression performance, we highlight the critical influence of actuator delay by contrasting policies trained with and without delay consideration, underscoring its significance for reliable flow control. Collectively, these contributions advance learning-based flow control by demonstrating the feasibility of translating DRL policies to real-time experimental hardware in high Reynolds number VIV applications.

The remainder of this paper is organized as follows. Section~\ref{sec:problem} describes the problem statement. Section \ref{sec:method} illustrates the experimental setup, including the fluid–structure apparatus, sensor, actuation system, and validation against published results. Section \ref{sec:DRL} details the deep reinforcement learning framework, including the state and action representations, reward formulation, and training procedure. Section \ref{sec:results} presents the experimental findings of the implementation of the DRL. Finally, Section \ref{sec:conclusions} summarizes the key outcomes and outlines directions for future research in experimental DRL-based flow control.

\section{\label{sec:problem}Problem Statement}
\begin{figure}
    \centering
    \includegraphics[width=0.6\linewidth]{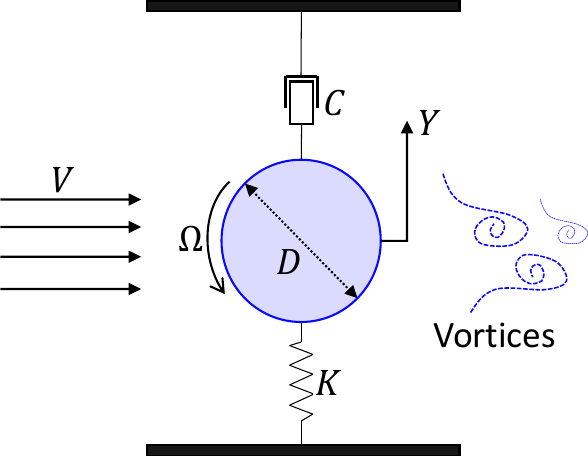}
\caption{A schematic representation of a circular cylinder in cross-flow. For illustration purposes, the dynamics of the cylinder is lumped into a single-degree-of-freedom model involving a lumped mass, $M$, stiffness, $K$, and damping coefficient, $C$.} 
\label{schematic diagram}
\end{figure}

We consider the system illustrated in Fig.~\ref{schematic diagram}, which consists of a cylinder of diameter, \( D \), and mass, \( M \), elastically mounted on a spring with stiffness, \( K \). The cylinder is free to oscillate transversely, i.e., in a direction perpendicular to a uniform incoming flow with density, \( \rho \), and steady velocity, \( V \). For certain values of \( V \), vortices are shed from the cylinder at a frequency near the natural frequency of the system, \( f_n = (1/2\pi)\sqrt{K/M} \). This frequency matching leads to resonant pressure forces that drive the cylinder into large-amplitude oscillations, denoted by \( Y(t) \), a phenomenon commonly referred to in the literature as vortex-induced vibration (VIV).

Our goal here is to use DRL to discover a control policy that imposes a rotation $\Omega(t)$ (recall Equation (\ref{Omega}))  on the cylinder such that the amplitude of the VIV is suppressed. This will be realized in an experimental setting using only sensor measurements of $Y(t)$ and its derivatives, and the actuation force itself when needed. The DRL has no knowledge of the flow parameters, or the lift forces which are a complex function of time, $t$, $\Omega (t)$, $Y(t)$ and its derivatives.

\section{\label{sec:method}Experimental System}
To achieve the goal of this study, an experimental system was created to mimic the conditions stated in the problem statement. The setup is shown in Fig.~\ref{setup1} and consists of four main components: $(i)$ the test cylinder and the mounting mechanism, $ii)$ the actuation mechanism, $iii)$ the measurement system, and the $iv)$ testing environment.

\subsection{Test cylinder and its mounting mechanism}
\label{VIVCylinder}
\begin{figure}[tb]
    \centering
    \includegraphics[width=0.75\linewidth]{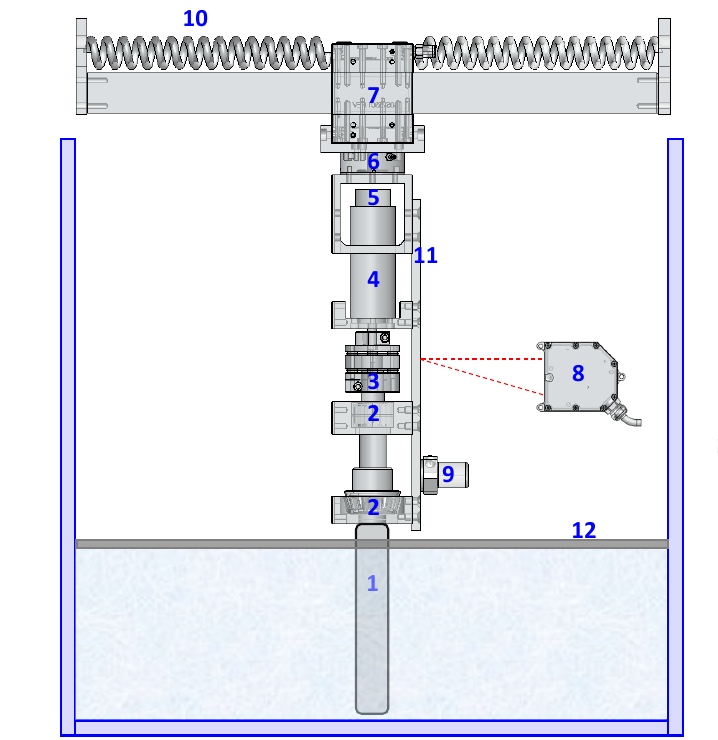}
    \caption{Schematic of the experimental setup with numbered components: \textcolor{blue}{1} Test cylinder, 
                \textcolor{blue}{2} Precision rotary bearings, 
                \textcolor{blue}{3} Flexible coupler, 
                \textcolor{blue}{4} DC motor, 
                \textcolor{blue}{5} Rotary encoder, 
                \textcolor{blue}{6} Force sensor controller, 
                \textcolor{blue}{7} Linear air bearing, 
                \textcolor{blue}{8} Laser displacement sensor, 
                \textcolor{blue}{9} Accelerometer, 
                \textcolor{blue}{10} Linear springs, 
                \textcolor{blue}{11}, Precision connecting rod.}
    \label{setup1}
\end{figure}

The test cylinder denoted by \textcolor{blue}{1} in Fig.~\ref{setup1} is machined from Aluminum and has a diameter $D = 17.5 \pm 0.01~\si{mm}$ and an immersed length of $L = \SI{160}{mm}$, yielding an aspect ratio $L/D \approx 9$. The cylinder is free on one end (the immersed end) and mounted on the other end onto a custom housing using two precision rotary bearings ``\textcolor{blue}{2}'' as shown in the figure. The rotary bearings, are rigidly mounted to a custom housing using alignment rods, which carry the weight and fix the cylinder along a defined rotational axis. The mounted end of the test cylinder is connected to a motor ``\textcolor{blue}{4}'' using a flexible coupler ``\textcolor{blue}{3}'', which mitigates any residual misalignment between the motor shaft and the rotating cylinder. The whole system including the test cylinder, the rotary motor, the rotary bearings and the housing 
are mounted on a linear track using a linear air bearing ``\textcolor{blue}{7}''. The air-bearing ensures low structural damping while constraining the motion solely in the cross-flow direction.  The air-bearing is connected on either side to two linear springs ``\textcolor{blue}{10}'' that provided the elasticity necessary to incite VIV.

To minimize end effects and encourage uniform vortex shedding, the gap between the cylinder's end and the platform is set to approximately $0.02D$ (i.e., \SI{1}{mm}), consistent with the setups used by Khalak \& Williamson~\cite{khalak1997investigation}. The total mass of the oscillating system, encompassing the cylinder, motor assembly, and movable parts of the air-bearing system, is $M = \SI{1095}{g}$. The displaced fluid mass is $M_d = \SI{360.07}{g}$, resulting in a mass ratio $m = M/M_d = 30.35$. Free-decay tests conducted in both air and still water were used to determine the natural frequencies and structural damping. The natural frequency in quiescent water was measured at $f_{n} = (1/2\pi) \sqrt{K/M} = \SI{1.96}{Hz}$, and the structural damping ratio in air was found to be $\zeta_{\text{air}} = C/(2\sqrt{KM}) \approx 1.02 \times 10^{-2}$.

\subsection{Actuation mechanism}
The cylinder is actuated to rotate around its minor axis using a brushed DC motor with a rated torque of 0.24 kg-cm. The DC motor is actuated using a DCC1002 motor controller (Phidget). The controller drives the motor using a pulse-width modulation (PWM) signal, where the duty cycle determines the applied voltage and thereby regulates the motor speed and direction. For system integration, the controller is interfaced through the Phidget library in Python, which provides a dedicated class straightforward instantiation of motor objects, parameter configuration, and execution of control commands.

\subsection{Measurement system} 
The measurement system is designed to capture the dynamic response of the rotating cylinder with high accuracy. Transverse body displacement was recorded using a laser range sensor (Micro-Epsilon) denoted by ``\textcolor{blue}{8}'' in Fig.~\ref{setup1} with a measurement range of ±0.1 mm. The digital output of the sensor is converted into an analog voltage signal via a C-Box signal converter, making it compatible with the acquisition system. Local acceleration of the body is measured using a piezoelectric accelerometer, denoted by ``\textcolor{blue}{9}'' in Fig.~\ref{setup1}, which is connected to a Dytran signal conditioning unit for filtering, thereby producing an analog voltage output. Rotational motion is monitored using a digital rotary encoder (E5-1000, US Digital, USA), whose quadrature pulses are converted into an analog signal before acquisition (denoted by ``\textcolor{blue}{5}'' in Fig.~\ref{setup1}). All measurement signals are thus available in analog form and are routed to NI6281 data acquisition board (National instruments). Finally, the NI DAQ board is interfaced with Python through a dedicated class, allowing motor actuation and sensing to be integrated and executed within a single environment. This platform  serves as the agent in the DRL framework, directly interacting with the experimental system by executing actions and receiving feedback for learning.

\begin{figure}[tb]
    \centering
    \includegraphics[width=0.75\linewidth]{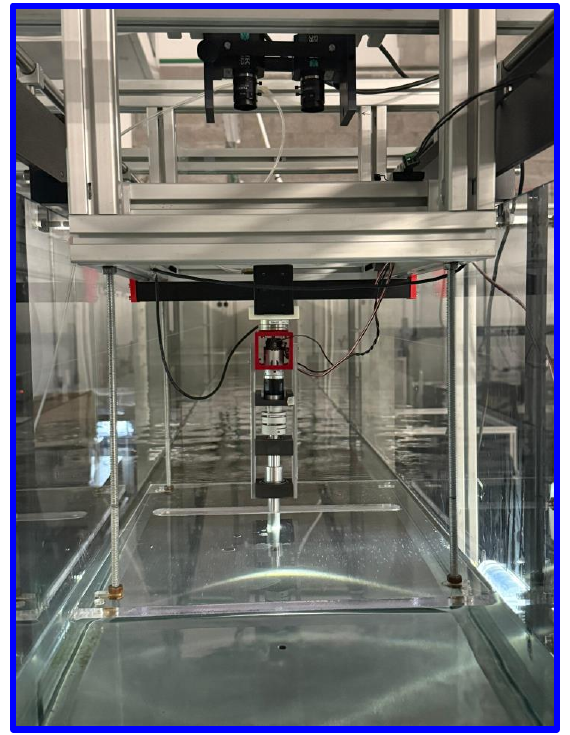}
    \caption{Experimental setup of the rotating cylinder within the flume. Key components of the EduPIV system—including the LED illumination, high-speed camera, and plexiglass flow stabilizer are also shown.}
    \label{setup2}
\end{figure}

The EduPIV system (Dantec Dynamics) is employed to capture the flow structures in the near wake of the rotating cylinder. The flow is seeded with polyamic tracer particles of nominal diameter $50~\mu$m, which are illuminated by a high-intensity LED light source. Imaging is performed using a high-speed camera with a resolution of $1960~\times~1280$ pixels. To mitigate disturbances from the free surface and ensure image clarity, a plexiglass sheet is placed above the measurement section to stabilize the flow. See Fig.~\ref{setup2} for details of the PIV system. Images are acquired at a frame rate of 310 frames per second. For each measurement, 1000 images are recorded. Image pre-processing and cross-correlation are performed using Dantec Dynamics software, providing vorticity contours of the wake downstream the rotating cylinder.

\subsection{Testing environment} 
All tests are performed in the recirculating free-surface water channel at the Laboratory for Applied Nonlinear Dynamics (LAND), see Fig.~\ref{setup2}. The test section of the water channel measures \SI{400}{mm} in width, \SI{500}{mm} in depth, and \SI{6000}{mm} in length. The free-stream velocity can be continuously varied within the range $0.056 \le V \leq 0.45~\si{m/s}$, and the turbulence intensity in the free stream is maintained below 1\%.

\subsection{Experimental verification}
\begin{figure}[tb]
    \centering
\includegraphics[width=0.95\linewidth]{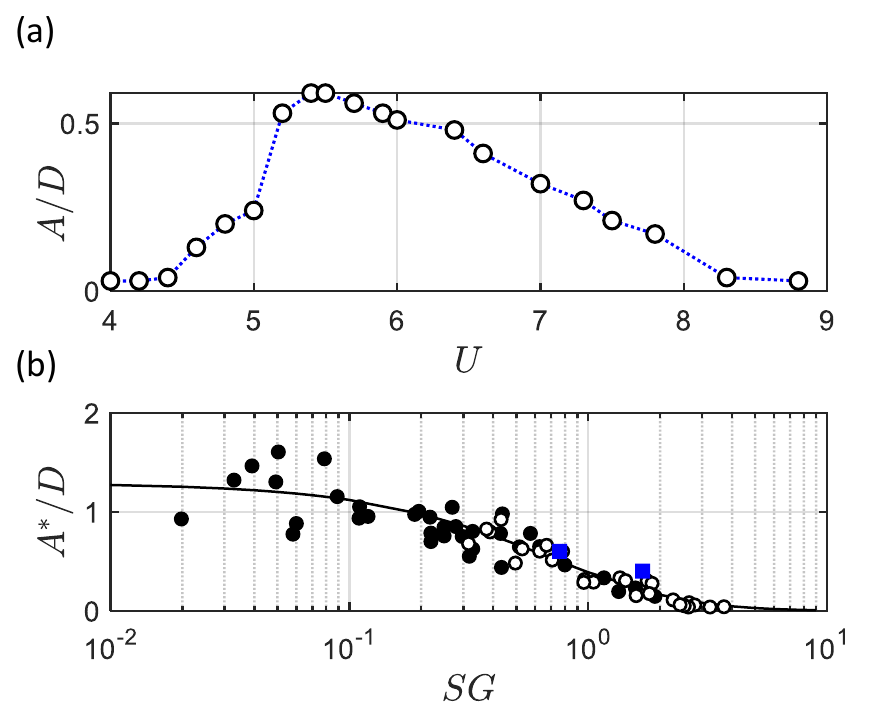}
\caption{(a) Variation of the normalized steady-state VIV amplitude, $A/D$, with the reduced velocity, $U$, highlighting the typical \textit{lock-in} curve. (b) Effect of the Skop-Griffin number, $SG$, on the peak response amplitude, $A^*/D$, within the lock-in region. Square (blue) markers denote the present experimental results. The solid line corresponds to the model of Griffin and Ramberg~\cite{griffin1976vortex}, while the solid and hollow circles represent data obtained from Paidoussis et al.~\cite{paidoussis2010fluid} in water and air, respectively.}  
\label{amplitudes}
\end{figure}

Before we delve into the design of the DRL controller, we verify that the experimental system described in section~\ref{VIVCylinder} does indeed undergo VIV in the absence of a prescribed rotary actuation. To this end, we first study variation of the dimensionless steady-state amplitude, $A/D$, of the cylinder as the reduced flow speed, $U= V/(f_n D)$ is varied. It can be clearly seen in Fig.~\ref{amplitudes}(a) that the steady-state response exhibits the typical \textit{lock-in} phenomenon, where large-amplitude responses occur over a range of reduced wind speeds that correspond to vortex shedding frequencies close to the natural frequency of the oscillator. The response clearly exhibits the typical characteristics of the \textit{lock-in} phenomenon with two distinct branches of solution: the initial branch, which exists for \( 4.5 \leq U \leq 5.1 \), the lower branch , which covers the range \( 5.1 \leq U \leq 9.0 \). The peak amplitude response is observed to reach \( A/D \approx 0.6 \) in the present experiments, compared with \( A/D \approx 0.95 \) in Khalak and Williamson~\cite{khalak1997investigation} and \( A/D \approx 1.0 \) in Zhao et al.~\cite{zhao2014three}. The reduction in the peak steady-state amplitude is expected considering the influence of the mass ratio, which is inherently higher in this setup due to the additional weight from the motor assembly.

As an additional confirmation of our findings that the cylinder oscillations are indeed due to VIV, we plot the peak amplitude $A^*/D$ of the steady-state response against the Skop-Griffin number, $SG$, of the system defined as \cite{skop1973,griffin1980vortex}
\begin{equation}    
SG = 2\pi^3 \text{St}^2\left(1+ m\right)\zeta,
\end{equation}
where $St \approx 0.21$ is the Strouhal number of the cylinder at large Reynolds numbers, $m=30.3$ is the mass ratio of the cylinder and $\zeta=0.012$ is the damping ratio of the system. As shown in Fig.~\ref{amplitudes}(b), we can clearly see that the coordinate of the resulting point falls nicely on the Skop-Griffin universal curve further confirming that the oscillations are indeed due to VIV. The process is repeated for a different cylinder with a mass ratio $m=40.1$ and a damping ratio $\zeta=0.018$. Again, the coordinate of the resulting point falls on the Skop-Griffin universal curve.

\section{\label{sec:DRL}Deep Reinforcement Learning}
Unlike traditional active control methods in fluid mechanics, which typically rely on reduced-order models or prior knowledge of the governing equations, deep reinforcement learning (DRL) enables the design of controllers without requiring an explicit model of the flow physics. In this study, the agent, implemented as a computer program, interacts iteratively with the experimental setup and adapts its control strategy to minimize vibrations. The DRL approach adopted here is based on proximal policy optimization (PPO)~\cite{ppo}.

The PPO agent samples a sequence of states, $s$, actions, $a$ and rewards, $r$, represented by $\eta$ defined as
\begin{equation}
\eta = (s_1, a_1, r_1), (s_2, a_2, r_2), \dots, (s_{N}, a_{N}, r_{N}),
\end{equation}
where $N$ is the number of times the DRL agent applies a policy $\pi_\theta(a_t, s_t)$ in each episode. Then, the agent pauses learning and updates the policy after each trajectory using the accumulated experience. The state space consists of the kinematic variables of the oscillating cylinder, represented by $Y/D$ and $\dot{Y}/f_nD$. The action space consists of one action corresponding to the voltage duty cycle applied to the motor controller which can take a value between [-0.4, 0.4]. The reward aims to mitigate the cylinder displacement (vibrations) and is defined as
\begin{equation}
r = -|Y/D|.
\end{equation}

\begin{figure}[tb]
    \centering
\includegraphics[width=0.85\linewidth]{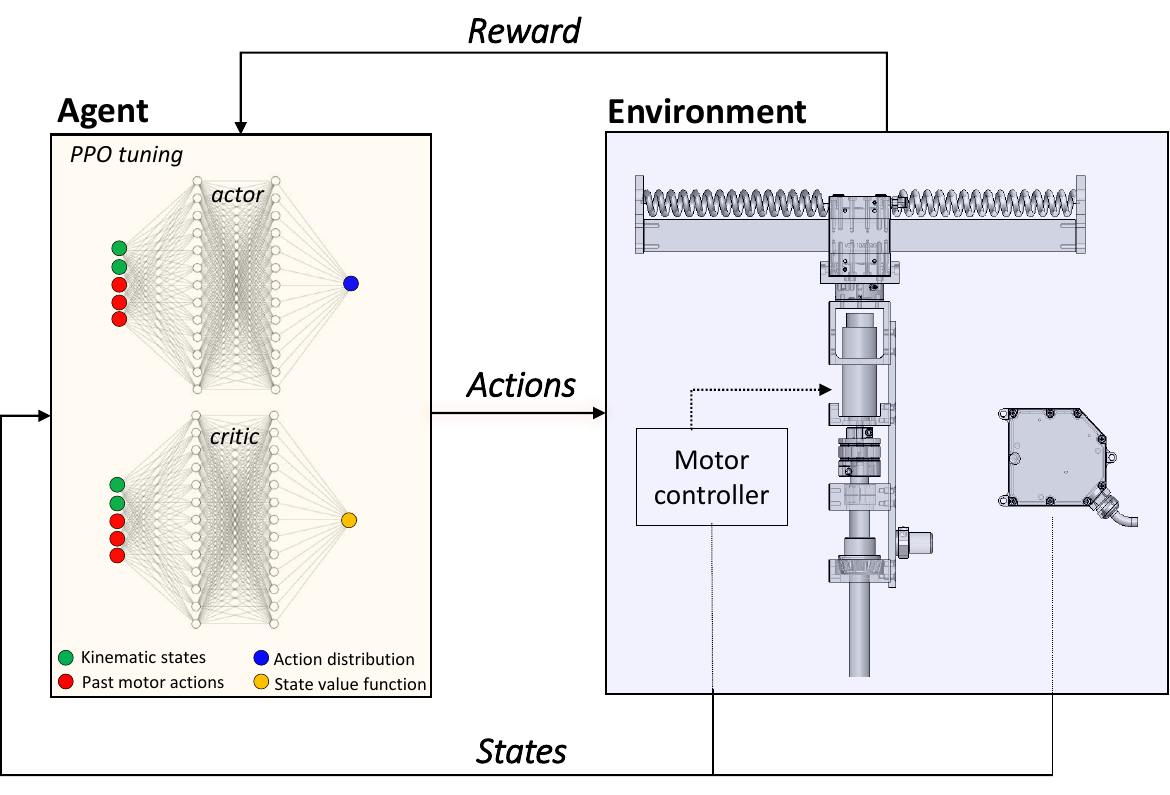}
\caption{Illustration of the deep reinforcement learning framework. The environment is coupled with the learning agent}
    \label{DRL}
\end{figure}

Figure~\ref{DRL} presents the network architectures of the standard PPO method. The algorithm comprises two neural networks: the \emph{actor} and the \emph{critic}. Both networks receive the same state vector as input and share a similar architecture, consisting of an input layer, two fully connected hidden layers with 64 units each, and an output layer. The critic network outputs an estimate of the state value function $V^\pi(s)$, while the actor network outputs the agent's policy $\pi_\theta$ distribution, i.e. the control strategy. More details on this framework can be found in~\cite{ppo}.


Deploying DRL in real-time experiments introduces challenges that are absent in simulations. First, the motor controller hardware operates with a minimum command update interval of \SI{100}{ms}, which imposes an upper limit on the control frequency. To accommodate this limitation, we deliberately select a system operating frequency around 1.96 Hz, which corresponds to approximately five control actions per oscillation period. Second, the motor exhibits an actuation delay. Preliminary tests showed a lag of roughly \SI{200}{ms} between the application of a voltage input and the attainment of its maximum rotational speed. Such delays may complicate reinforcement learning because the executed command is not immediately reflected in the observed state. A widely used strategy in DRL to address this issue is to augment the state representation with a history of past control inputs \cite{weissenbacher2025reinforcement}. Given that the motor requires about \SI{200}{ms} (i.e., two sampling intervals) to reach full speed, we include the two most recent control actions in the state vector. This ensures that the policy can correctly associate the current flow state with the cumulative effect of prior actions, effectively embedding the delayed actuation dynamics into the training framework.

In the present study, the duration of each training episode is set to \(T = 25T_0\), where \(T_0\) is the natural period of the cylinder in the uncontrolled case. With \(T_0 \approx 0.5\ \si{s}\), the total episode duration is \(T \approx 12.5\ \si{s}\). Within each episode, the flow states and control actions are updated 128 times. After training converges, we switch to a deterministic control phase in which the neural network weights remain fixed. During this stage, the actor network alone is used to map the observed state to the corresponding control action. The deterministic control is conducted for \SI{50}{s}, with the same actuation frequency as in the training.

\section{\label{sec:results}Results and Discussion}
This section presents the experimental findings of the DRL-based rotary control framework. Here, we assess the performance of the DRL strategy when implemented using only state feedback and compare it to the well-known open-loop \textit{lock-on} sinusoidal control strategy. We examine the effect of actuator delays within the control loop on the overall performance of the DRL control policy and discuss an approach to account for them.

\subsection{DRL control policy}
\begin{figure}[tb]
    \centering
    \includegraphics[width=0.95\linewidth]{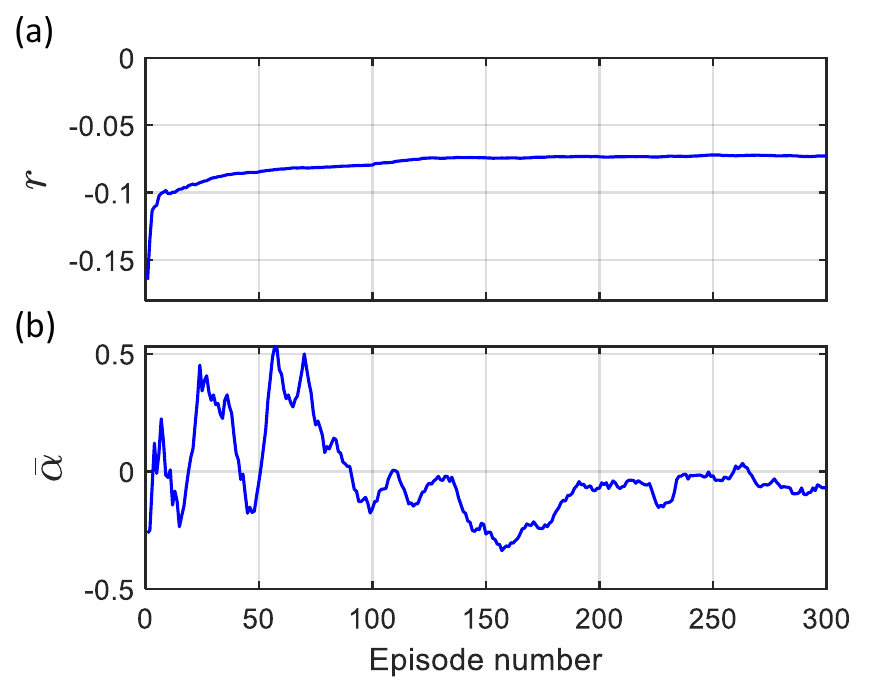}
\caption{(a) Variation of the average reward, $r$, with the number of episodes. (b) Variation of the mean rotational speed, $\bar{\alpha}$, with the number of episodes}
    \label{direct_reward}
\end{figure}

We first implement the DRL strategy on the experimental system assuming that the control action can be applied instantly without any actuator lag.  In such a scenario, the state, action, and reward are directly coupled within each control cycle. Figure~\ref{direct_reward}(a) shows how the training reward develops over successive episodes. The reward initially takes on negative values, then rapidly increases, and eventually stabilizes, indicating that the DRL agent has converged and established a stable control policy. Convergence is observed after approximately 300 episodes, which corresponds to around 60 minutes of training.

Evolution of the cylinder's normalized mean rotational speed defined as $\bar{\alpha} = \bar{\Omega}_0 D/(2U)$ ($\bar{\Omega}_0$ is the dimensional mean rotational speed), throughout the training process is shown in Fig.~\ref{direct_reward}(b). At the onset of training, $\bar{\alpha}$ deviates from zero, reflecting a directional bias in the initial control input. As training progresses however, the mean rotational speed gradually converges toward zero, indicating that the learned policy adopts a symmetric back-and-forth rotation of the cylinder. This oscillatory motion is crucial for maintaining the cylinder displacement near $Y = 0$, thereby stabilizing the structure and mitigating the VIV. 

\begin{figure}[tb]
    \centering
    \includegraphics[width=\linewidth]{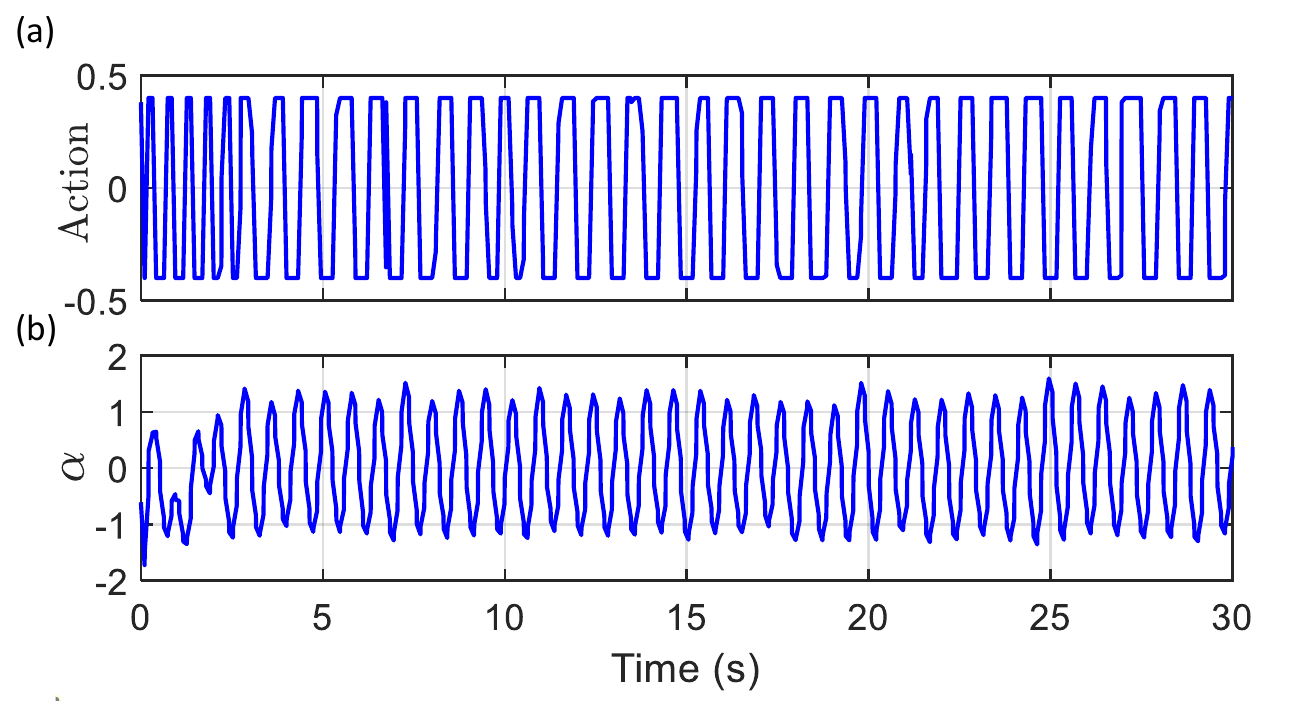}
\caption{Results of the DRL-based deterministic control. (a) Control action prescribed by the agent, represented as the applied voltage duty cycle. (b) Resulting normalized motor rotational speed, $\alpha$, as measured by the rotary encoder.}
    \label{direct_signal}
\end{figure}

Upon convergence of the DRL training, the control policy learned by the ``actor'' network is extracted. A deterministic control experiment is then performed for a duration of $t = 30$ s. Figure~\ref{direct_signal} illustrates the control actions learned by the agent along with the corresponding normalized rotational speed, $\alpha (t)=\Omega (t) D/(2U)$, of the cylinder. We observe that the motor actuates in discrete steps, yet the resulting rotational speed exhibits a smooth sinusoidal profile. This indicates that the DRL agent successfully approximated the nonlinear mapping between the voltage duty cycle of the motor and its resulting velocity through interaction with the environment without the need for complex system identification and transfer function modeling; a step that is necessary in conventional control design.

\begin{figure}
    \centering
    \includegraphics[width=\linewidth]{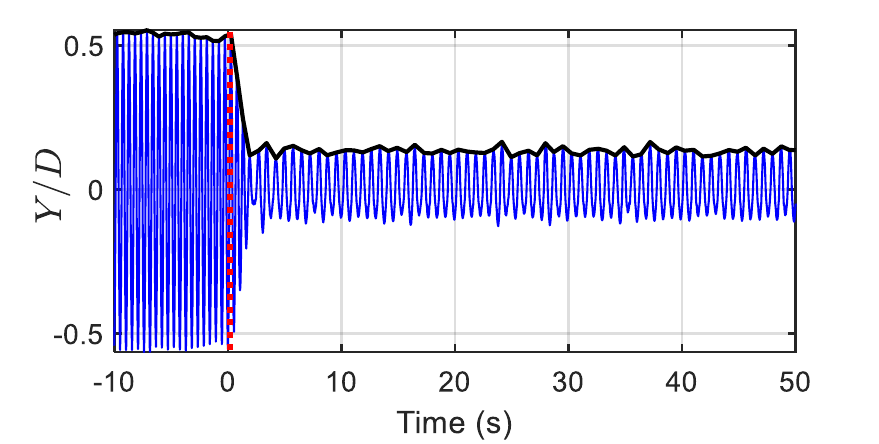}
\caption{Time history of the non-dimensional displacement, $Y/D$. Red-dashed line marks the onset of actuation.}
    \label{direct_result}
\end{figure}

A closer examination of Fig.~\ref{direct_signal}(a) reveals that the learned control policy operates in two distinct phases characterized by different actuation frequencies. Initially, the agent applies high-frequency periodic inputs, which gradually transition to a lower frequency over time. This behavior suggests that the agent prioritizes a rapid suppression of oscillations at the onset, before settling into a more stable, sustained control regime. The corresponding normalized time history of oscillations, $Y/D$, is presented in Fig.~\ref{direct_result}. It is clear that, once the control is activated (indicated by the dashed red line), the amplitude of VIV quickly drops to significantly lower levels, highlighting the impact of the initial high-frequency actuation. Over time, the response stabilizes at a lower frequency, below the natural frequency of the system. The post-control oscillation amplitude is approximately 80\% of the uncontrolled case, highlighting the effectiveness of the learned control policy.

\subsection{Rotary sinusoidal lock-on control}
\begin{figure}[tb]
    \centering
    \includegraphics[width=0.75\linewidth]{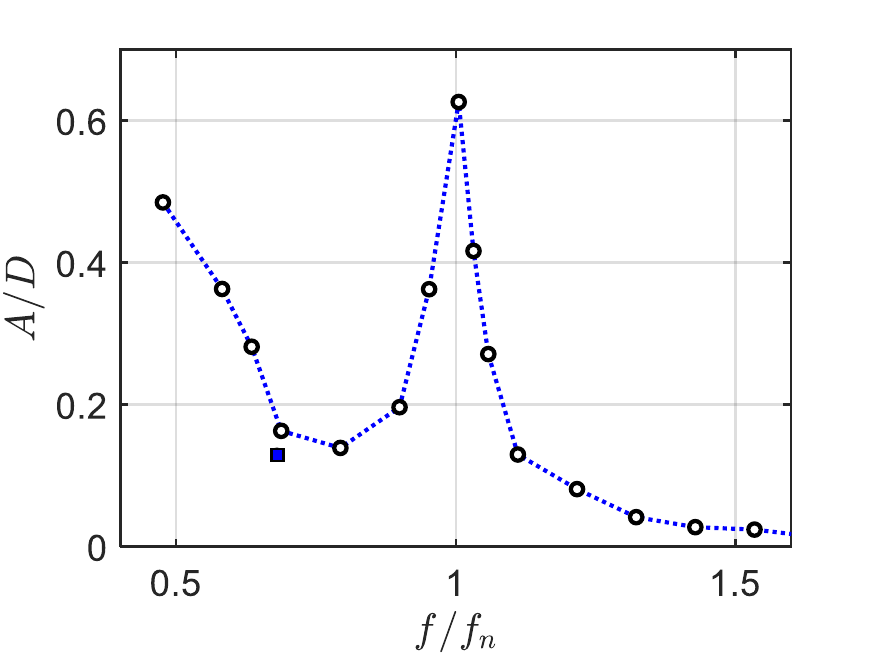}
\caption{Amplitude response of the cylinder under sinusoidal actuation at varying frequency ratios. Square marker shows a comparison against the DRL-based control strategy}
    \label{PID}
\end{figure}

Given that the learned DRL policy produces oscillatory control signals, it is natural to benchmark its performance against a the traditional rotary sinusoidal \textit{lock-on} control algorithm. To this end, a PID-based controller is implemented to actuate the cylinder to rotate sinusoidally at a normalized speed $\alpha(t)=\alpha_0 \cos (2 \pi f_r t)$. The magnitude $\alpha_0$ of the sinusoidal speed was kept constant at one to ensure consistency with the peak amplitude achieved by the DRL policy (see Fig.~\ref{direct_signal}(b)). The frequency $f_r$ of the rotational speed is varied in the range $0.4 f_n \leq f_r \leq 1.6 f_n$, where $f_n$ is the natural frequency of the oscillating cylinder. The resulting magnitude of the steady-state VIV, $A$, normalized by the cylinder diameter, $D$, is recorded and plotted against $f_r/f_n$ as shown in Fig. \ref{PID}.

It is observed that as the normalized forcing frequency \( f_r/f_n \) increases from 0.4 to 0.8, the body begins to synchronize with the forcing frequency, resulting in a sharp reduction in the amplitude of the VIV response. This trend continues up to \( f_r/f_n = 0.8 \). Beyond this point, the amplitude rises rapidly with increasing \( f_r/f_n \), reaching a peak value of \( A/D = 0.65 \) at \( f_r/f_n = 1 \). As the forcing frequency increases further, the steady-state amplitude of the VIV drops sharply, eventually falling below \( A/D = 0.05 \) at \( f_r/f_n = 1.6 \).

To evaluate the performance of the DRL policy in comparison to sinusoidal rotary forcing, a Fast Fourier Transform (FFT) is applied to the signal shown in Fig.~\ref{direct_signal}(b) to determine the dominant frequency \( f_r \) of the rotational motion produced by the DRL controller. The corresponding point \((f_r/f_n, A/D)\) is plotted as a blue circle on the frequency–response curve in Fig.~\ref{PID}. This point closely follows the sinusoidal response trend and aligns with the onset of the lock-on region, suggesting that the learned control strategy takes advantage of the lock-on mechanism, where vortex shedding synchronizes with the forcing frequency. 

\begin{figure}[tb]
    \centering
    \includegraphics[width=0.7\linewidth]{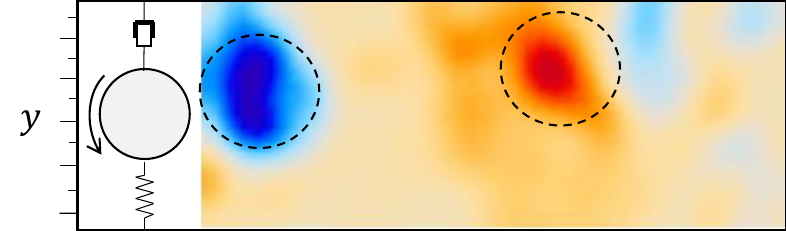}
\caption{Vorticity contours under the DRL-based control.}
    \label{wakePIVLowFreq}
\end{figure}

Additional evidence of this behavior is provided by PIV measurements of the wake behind the cylinder. As shown in Fig.~\ref{wakePIVLowFreq}, the vorticity contours display a wake pattern consisting of two oppositely-signed vortices shed per oscillation cycle, consistent with the classical 2S mode described in the literature~\cite{Williamson2004}. The shedding frequency matches the actuation frequency, indicating that the DRL-based controller effectively modifies the wake dynamics to achieve \textit{lock-on} synchronization, thereby reducing the vibration amplitude.

\subsection{Impact of actuator dynamics on DRL control}
Direct implementation of the DRL policy using only state feedback results in clear attenuation of VIV, achieving approximately 80\% suppression. However, this level of performance remains suboptimal when compared to the nearly complete suppression achieved by the sinusoidal rotary control strategy at higher frequency ratios (see Fig.~\ref{PID}). This raises the question: why does the DRL controller converge toward a low-frequency actuation policy, rather than adopting a high-frequency excitation that yields greater attenuation? We hypothesize that this discrepancy stems from actuator dynamics; specifically, delays in the motor response, which may hinder the agent’s ability to effectively implement high-frequency control actions.

\begin{figure}[tb]
    \centering
    \includegraphics[width=0.75\linewidth]{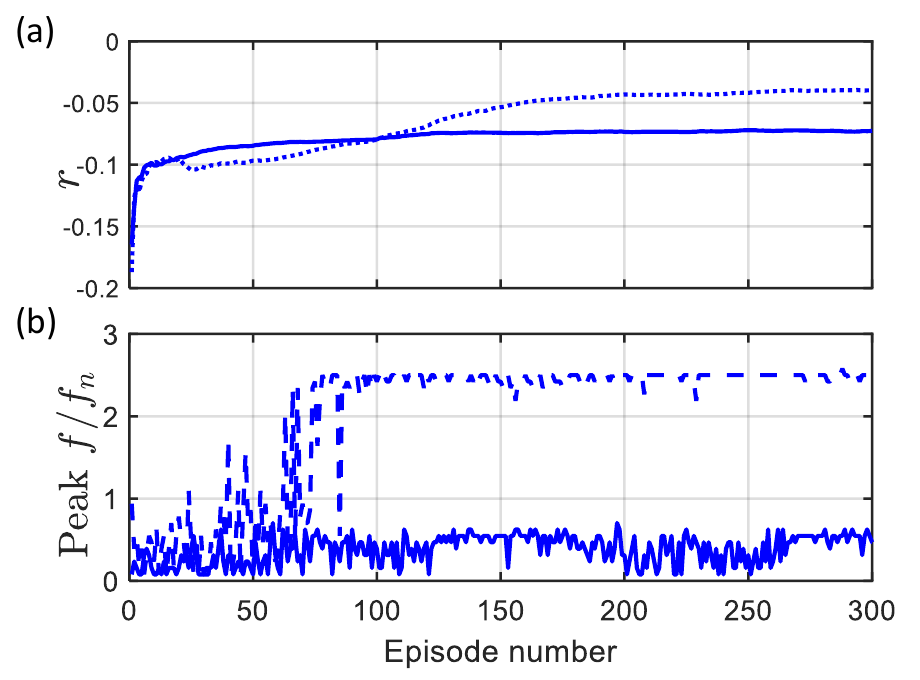}
\caption{(a) Evaluation of the reward without augmenting past actions (solid line), and when augmenting two past actions in the state vector (dashed lines). (b) Variation of the dominant frequency of the rotary forcing with the episode number. No past actions (solid line), and two past actions (dashed lines).}
\label{delay_rewards}
\end{figure}

To address this limitation, the state vector is augmented with past motor actions, as described in Sec.~\ref{sec:DRL}. Training performance is evaluated for two cases: one without past actions (\( n = 0 \)) and one with two past actions included in the state vector (\( n = 2 \)). The corresponding reward histories are shown in Fig.~\ref{delay_rewards}(a), where the solid line represents the case without past actions, and the dashed line corresponds to the augmented state. 

It is evident that excluding past actions leads to lower rewards, indicating reduced capability in attenuating vibrations. In contrast, including two previous motor commands allows the agent to achieve consistently higher rewards, demonstrating improved VIV suppression performance.

To compare the characteristics of the control input with and without the inclusion of past actions, we analyze the dominant frequency component of the rotational speed throughout training for both cases, as shown in Fig.~\ref{delay_rewards}(b). Initially, both approaches exhibit high variance in the dominant frequency, reflecting the agent's exploration during early training stages. As training progresses, the variance decreases and the dominant frequencies begin to diverge. Without access to past actions, the dominant frequency ratio remains below 1. In contrast, when past motor actions are incorporated into the state, the dominant frequency ratio shifts to significantly higher values, exceeding \( f_r/f_n = 2.5 \). This indicates that including past actions enables the agent to explore and exploit higher-frequency control strategies, which correspond to substantially increased rewards, as shown in Fig.~\ref{delay_rewards}(a). 

\begin{figure}[tb]
    \centering
    \includegraphics[width=\linewidth]{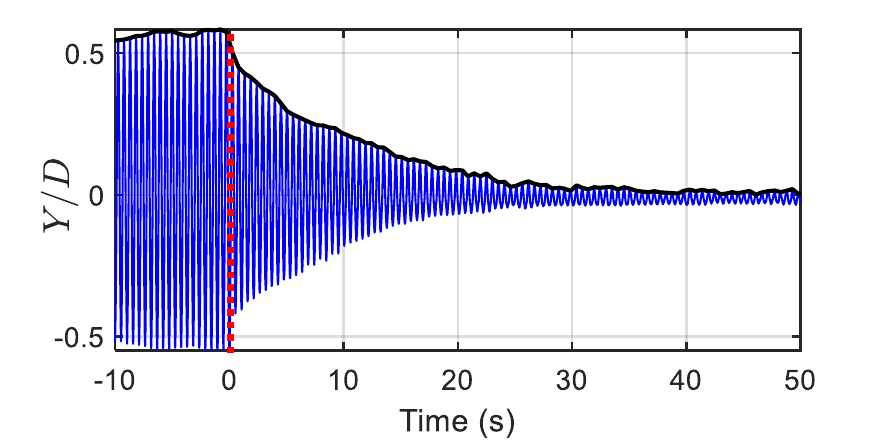}
    \caption{Time history of the non-dimensional displacement, $Y/D$, with past motor commands included in the state vector. Red-dashed line marks the onset of actuation.}
    \label{delay_result}
\end{figure}

The learned control policy, which includes past actions, is deployed to attenuate the VIV of the cylinder by allowing the actor network to apply deterministic control over a 60 second duration. The time history of the resulting VIV response, shown in Fig.~\ref{delay_result}, clearly demonstrates that accounting for past actions leads to enhanced performance as the DRL agent is now able to achieve more than a 95\% reduction in oscillation amplitude.

The corresponding wake vorticity contours reveal distinct characteristics. While the vortex shedding remains in a 2S mode, the vortices are now shed into two narrowly spaced rows due to the enhanced vibration suppression. This wake pattern closely follows mode I observed for a rigidly mounted cylinder at varying frequency ratios, as reported by Choi et al.~\cite{Choi2002}.
\begin{figure}[tb]
    \centering
    \includegraphics[width=0.75\linewidth]{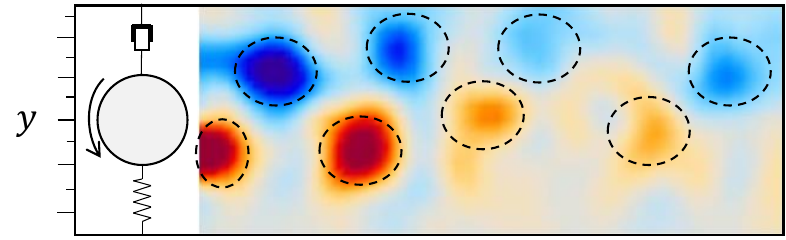}
    \caption{Vorticity contours under the DRL-based control with past motor commands included in the state vector.}
    \label{vorticity_delay_result}
\end{figure}

The contrasting performance between the two strategies arises from how the agent interprets environmental information. In the first case, without memory, each state observation is linked only to its simultaneous action. Lacking the ability to explicitly model actuation delays, the policy converges to a slow, more conservative approach, producing low-frequency control inputs.  In contrast, augmenting the input space with past actions provides the agent with an implicit memory, enabling it to infer the system’s delayed response dynamics. This enhanced representation allows the policy to generate higher frequency control inputs that more effectively suppresses vibration.

\subsection{Comparison to a recent numerical study}
An interesting distinction arises when comparing the present experimental results with recent numerical DRL studies. In our experiments, the learned controller suppresses VIV primarily through the \textit{lock-on} mechanism, in which the actuation modifies the vortex shedding frequency and synchronizes the structural response with the forcing. In contrast, Ren \textit{et al.}~\cite{ren2024deep} reported that their DRL-guided rotary control achieved vibration suppression without altering the shedding frequency. Instead , the DRL policy stabilizes the flow by driving the growth rates of dominant modes to negative values. Several factors may account for this discrepancy. First, differences in Reynolds number (~\cite{ren2024deep} simulated $Re=100$ while our experiment is at $Re=3000$) may influence the instability dynamics and thus the control pathways available to the DRL agent. Second, the numerical setup eliminates experimental complexities such as measurement noise or actuator delays, which are intrinsic to physical systems and may favor \textit{lock-on} type strategies. Finally, the higher sampling and action frequencies employed in the simulations enable more rapid policy updates, which are constrained in our experimental study. These observations suggest that DRL can converge to distinct yet effective control strategies, highlighting a promising direction for future research in VIV control.

\section{\label{sec:conclusions}Conclusions}
This study presents an experimental implementation of deep reinforcement learning (DRL) for active flow control (AFC) of vortex-induced vibrations (VIV) in a circular cylinder at high Reynolds numbers ($Re = 3000$) using rotary control. Unlike previous studies which are numerical in nature and conducted at low Reynolds numbers, the present work demonstrates, for the first time, the successful deployment of DRL policies for real-time experimental control of VIV at a high Reynolds number, accounting for practical challenges such as actuation delays and limited sampling rates.

The key findings of this study can be summarized as follows:
\begin{enumerate}
    
    \item When incorporating state feedback only (displacement and velocity of the oscillating cylinder) into the learning algorithm, the DRL agent is able to discover a low-frequency rotary control strategy that suppresses oscillations by up to 80\% relative to the uncontrolled case. The learned DRL policy resembles the well-known \textit{lock-on} phenomenon, which synchronizes vortex shedding with the actuation frequency. In the process, the DRL agent is able to directly map discrete voltage duty-cycle commands into rotary speed, thereby bypassing the need for an intermediate control model. 
    
   \item Due to actuator delays, the policy learned using state feedback alone was limited to low-frequency control strategies, resulting in suboptimal performance compared to that enabled by high-frequency rotary actuation exploiting the lock-on phenomenon. To overcome this limitation, the state feedback vector is augmented with previous control actions, allowing the DRL agent to account for actuator dynamics and temporal dependencies. This augmentation enabled the agent to discover higher-frequency actuation strategies, achieving over 95\% vibration suppression. These results highlight the critical role of temporal information in optimizing control performance under realistic actuation constraints.
\end{enumerate}

We plan to extend the experimental DRL framework to explore energy harvesting strategies in vortex-induced vibrations. At the same time, we aim to perform experiments at higher Reynolds numbers to approach more realistic flow conditions. These efforts are intended to bridge the gap between the powerful capabilities of machine learning and the practical implementations in active flow control.

\section*{acknowledgments}
The authors would like to thank Nikolas Giakoumidis (CAIR) for discussions on the setup design and Vijay Dhanvi (NYUAD Machine Shop) for apparatus fabrication.

\bibliographystyle{elsarticle-num}\biboptions{sort&compress}
\bibliography{mybib}   

%
%
\newpage

\renewcommand{\theequation}{\thesection.\arabic{equation}}
\renewcommand{\thefigure}{\thesection.\arabic{figure}}
\renewcommand{\thesection}{\Alph{section}}
\end{document}